\documentclass[runningheads]{llncs}
\usepackage{graphicx,url}
\usepackage{mathrsfs}  
\usepackage[utf8]{inputenc}  
\usepackage{algorithm}
\usepackage{algpseudocode}
\usepackage{dblfloatfix}
\usepackage{minibox}
\usepackage{color}
\usepackage{amssymb}
\begin{document}
\title{Modeling, comprehending and summarizing textual content by graphs\thanks{Supported by CAPES, CNPQ and FAPERGS.}}
%% gui: caso seja possível sugiro troca de título para: Modeling cross-domain graph-based summaries with redundancy control

% \titlerunning{Abbreviated paper title}
% If the paper title is too long for the running head, you can set
% an abbreviated paper title here

\author{
Vinicius Woloszyn\inst{1}\orcidID{0000-0003-3554-5580} \and
Guilherme Medeiros Machado\inst{1}\orcidID{0000-0001-5283-9228} \and
Leandro Krug Wives\inst{1}\orcidID{0000-0002-8391-446X} \and 
José Palazzo Moreira de Oliveira \inst{1}\orcidID{ 0000-0002-9166-8801}
}
\authorrunning{Wolozsyn et al.}
\institute{PPGC - Instituto de Informática - UFRGS, Porto Alegre RS, Brazil \\
\email{\{vwoloszyn, guimmachado, wives, palazzo\}@inf.ufrgs.br}\\
% \url{http://www.springer.com/gp/computer-science/lncs} 
}

\maketitle              % typeset the header of the contribution
\begin{abstract} %lwives: diminui, pois tem mais de 200 palavras!
Automatic Text Summarization strategies have been successfully employed to digest text collections and extract its essential content. Usually, summaries are generated using textual corpora that belongs to the same domain area where the summary will be used. Nonetheless, there are special cases where it is not found enough textual sources, and one possible alternative is to generate a summary from a different domain. %An example comes from the learning domain when it is necessary the generation of a movie summary (movie domain) to be used in the in a film-based lesson plan (learning domain).
One manner to summarize texts %(or sets of texts) 
consists in using a graph model. This model allows giving more importance %(weight) 
to words corresponding to the main concepts from the target domain found in the summarized text. This gives the reader an overview of the main text concepts as well as their relationships. However, this kind of summarization presents a significant number of repeated terms when compared to human-generated summaries. In this paper, we present an approach to produce graph-model extractive summaries of texts, meeting the target domain exigences and treating the terms repetition problem. To evaluate the proposition, we performed a series of experiments showing that the proposed approach statistically improves the performance of a model based on Graph Centrality, achieving better coverage, accuracy, and recall. % lwives: o foco aqui deveria ser a geração de grafos para compreender um texto e não melhorar a performance!!! vai acabar ficando igual aos anteriores? 
%lwives: está muito parecido com o encaminhado ao SBIE.. A ideia era focar em geração de grafos para representar melhor o conteúdo de um texto... 

\end{abstract}
\section{Introduction}
Automatic Text Summarization (ATS) systems play a significant role by extracting essential content from textual documents. This is important given the exponential growth of textual information. Despite not being one of the newest areas of research, there are still open-ended summarization issues that pose many challenges to the scientific community~\cite{saggion2013automatic}. One of the examples is when one summary must be generated prioritizing sentences that present terms of another specific domain (cross-domain summarization).%, such as the educational or the medical ones \cite{WoloszynBeatnik2017}.
Another example is the redundancy problem that occurs when a wrong text modeling leads to a repetition of content~\cite{saggion2013automatic}. %gui: o que é uma wrong text representation?

%The cross-domain summarization is a strategy to generate personalized summaries. The need for such personalization happens in situations when a specialized summary must be extracted from general purpose documents. For instance, if a professor wants to know better what are the educational aspects of a movie she is hoping to use in her class, and to do so, she is looking to other peoples' comments about that movie. Other example: imagine a person who is shopping a new novel hoping to find one that also presents, for instance, botanical knowledge during the story.

%vini:  
The cross-domain summarization is a strategy to generate biased summaries, which generally favors a subject. The need for such bias happens in situations when a summary containing specific aspects must be extracted from general purpose documents. For instance, if a teacher wants to know better what are the educational aspects of a movie she is hoping to use in her class, and to do so, she is looking to other peoples' comments about that movie. Another example: imagine a person who is shopping a new novel hoping to find one that also presents, for instance, historical facts of a city during the story.

%The redundancy problem on the other hand is a common problem that happens when a wrong text modeling can benefits the generation of summaries that repeat the most central sentences or select a set of very similar ones in the documents. This causes a gain in accuracy but generates redundant summaries with poor coverage of the text aspects.

%Most works on summarization rely on supervised algorithms such as classification and regression~\cite{Xiong2011,Zengwu2013,Yang2015}. However, the quality of results produced by supervised algorithms is dependent on the existence of a large, domain-dependent training dataset. One drawback of such strategy is that those datasets are not always available and their construction is labor-intense and error-prone since documents must be manually annotated to train the algorithms correctly. Thus, unsupervised methods are a relevant alternative to avoid these problems. 

Most works on summarization rely on supervised algorithms such as classification and regression~\cite{Xiong2011,Zengwu2013,Yang2015}. However, the quality of results produced by supervised algorithms is dependent on the existence of a large, domain-dependent training dataset. One drawback of such strategy is that those datasets are not always available and their construction is labor-intense and error-prone since documents must be manually annotated to train the algorithms correctly. 

Unsupervised models, conversely, are an interesting strategy for a situation where there are not enough textual sources since it does not need a large training set for the learning process. However, a common problem of these models is redundancy. It happens when a wrong text modeling can benefit from the generation of summaries that repeat the most central sentences or select a set of very similar ones in the documents. This causes a gain in accuracy but generates redundant summaries with poor coverage of text aspects.

To meet the cross-domain summarization needs and mitigate the redundancy problem, we propose an unsupervised graph-based model to generate cross-domain summaries. The generated graphs are able to uncover the main topics (concepts) of a document or a set of documents. To do so, the summarization algorithm focus on the most relevant, i.e., central, nodes using pre-determined domain corpora and nodes' relationships. In our experiments, this combination of cross-domain generation avoiding redundancy, improves Graph-Based ATS system's achieving better coverage, precision, and recall. 

%vini: acho que é muita informação para o leitor. Isso me parece ser mais trabalhos relacionados.
% Several unsupervised text summarization approaches have been successfully employed graph representations of documents, where the nodes represent sentences, and the edges account for the textual similarity between pairs of sentences  \cite{erkan2004lexrank,mihalcea2004textrank,tsur2009revrank,ganesan2010opinosis,wu2011unsupervised,woloszyn2017mrr,otterbacher2009biased}. The Graph algorithms (e.g., random walks, PageRank or HITS) then apply Graph Centrality to weight the relevance of sentences and make decisions about their pertinence and inclusion in the final summary. However, Graph Centrality strategies suffers from a related problem in which the larger groups tend to have higher scores, and in attempting to construct a maximum centrality list, the top items tend to belong to the same group, which favor the redundancy problem~\cite{everett2005extending}.

% Since redundancy or repetition is an unwanted behavior in ATS systems, the presented approach also perform a re-ranking of the initial centrality index to improve coverage and decrease redundancy. In our experiments, this combination of cross-domain generation avoiding redundancy, improves Graph-Based ATS system's achieving better coverage, precision, and recall. 

The contributions of this work are the following: 1) it is an unsupervised cross-domain summarization, i.e., it does not depend on specific annotated training set; 2) it address the redundancy problem performing a re-ranking of the initial centrality index to improve coverage and decrease redundancy; and 3) considering two distinct datasets, the results obtained in our experiments were significantly superior to the baselines analyzed. 

The rest of this paper is organized as follows. Section \ref{sec:background} presents the background of the area and related work. Section \ref{sec:model} details the proposed model. Section \ref{sec:case_study} provides a case study, and Section \ref{sec:experiment_design} describes the design of our experiments. Section \ref{sec:results} discusses the results. Section \ref{sec:conclusion} summarizes our conclusions and presents future research directions.
\section{Background}\label{sec:background}

% Automatic Text Summarization (ATS) techniques have been successfully employed on user-content to highlight the most relevant information among  documents~\cite{erkan2004lexrank,ganesan2010opinosis,saggion2013automatic,ramos2017experimental}. Regarding techniques usually employed, several works have explored unsupervised methods based on graph centrality. For instance, MRR \cite{woloszyn2017mrr}, which combines the centrality scores and explicit human feedback to produce a ranking of relevant documents. Another example is presented by Wu et al. 2011 \cite{wu2011unsupervised} that combines the centrality scores of each sentence with the documents' length to produce an overall centrality score for each review. However, this method does not scale well due to the granularity of centrality, which implies the dual use of PageRank and requires pre-processing to identify specific textual features. 

% There are also studies using supervised learning strategies to predict text relevance~\cite{Zengwu2013,Yang2015}. Additionally, the use of regression algorithms consistently improves the prediction of helpfulness~\cite{wan2013}. However, a typical drawback of supervised learning approaches is that the quality of results is heavily influenced by the availability of a large, domain-dependent annotated corpus, to train the representation model. In this sense, unsupervised learning techniques are attractive because they do not imply the cost of corpus annotation either training. 

Automatic Text Summarization (ATS) techniques have been successfully employed on user-content to highlight the most relevant information among  documents~\cite{erkan2004lexrank,ganesan2010opinosis,saggion2013automatic,ramos2017experimental}. Regarding techniques usually employed, several works have explored supervised learning strategies to predict text relevance~\cite{Zengwu2013,Yang2015}. Additionally, the use of regression algorithms consistently improves the prediction of helpfulness~\cite{wan2013}. However, these supervised techniques need annotated corpus for the training process, which for the most of the cross-domains cases is unavailable.

%Many unsupervised summarizers are based on graph centrality. For instance, MRR \cite{woloszyn2017mrr}, which combines the centrality scores and explicit human feedback to produce a ranking of relevant documents. Another example is presented by Wu et al. 2011 \cite{wu2011unsupervised} that combines the centrality scores of each sentence with the documents' length to produce an overall centrality score for each review. 

Graph Centrality is also widely employed on unsupervised extractive summarization systems where a graph representation of documents is used to weight sentences relevance on a set of documents~\cite{erkan2004lexrank,mihalcea2004textrank,woloszyn2017mrr,WoloszynBeatnik2017}. Based on that, central nodes indicate that the sentence they represent is relevant in the group of documents. Let $S$ be a set of all sentences extracted from the input document(s), a graph representation $G=(V,E)$ is created, where $V=S$ and $E$ is a set of edges that connect pairs $\langle u,v \rangle \in V$. Then, the score of each node is usually given by an algorithm like PageRank~\cite{page1999pagerank} or HITS~\cite{hits1999}. 

\begin{figure}[t]
     \centering
     \includegraphics[width=1\textwidth]{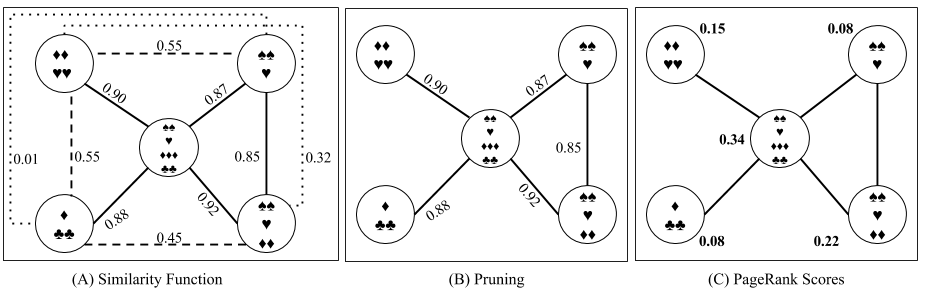}
 \caption{Illustration of Graph centrality steps, where symbols represent text words.}
 \label{fig:Repe_GC}
 \end{figure}

Figure \ref{fig:Repe_GC} depicts the general steps of a summarization system based on Graph Centrality: (a) it builds a similarity graph between pairs of sentences; (b) it prunes the graph by removing all edges that do not meet a minimum threshold of similarity; (c) it uses PageRank to calculate the centrality scores of each node. Then a Greedy strategy is employed, where the centrality index produces a ranking of vertices' importance, which is used to indicate the ranking of the most relevant sentences to compose the final summary. This a well-known strategy used as the basis for many novel unsupervised approaches~\cite{wu2011unsupervised,cheng2011relin,woloszyn2017mrr,al2017sentiment}. 

\textbf{PageRank}~\cite{page1999pagerank} computes the centrality of nodes, where each edge is considered as a vote to determine the overall centrality score of each node in a graph. However, as in many types of social networks, not all of the relationships are considered of equal importance. The premise underlying PageRank is that the importance of a node is measured in terms of both the number and the importance of vertices it is related to. The PageRank centrality function is given by:

 \begin{equation}\label{equationPR}
 PR(u)=\sum\limits_{v \in B_u} \frac{ PR(v) }{ N_v  }
 \end{equation}

 \noindent
where $B_u$ is the set containing all neighborhood of $u$ and $N_v$ the number of neighborhoods of $v$.

However, this strategy is normally employed with no restrictions that ensure an empty or a minimal intersection between sentences~\cite{erkan2004lexrank,wu2011unsupervised,cheng2011relin,woloszyn2017mrr,al2017sentiment}. This lack of restrictions would increase the overall redundancy on these approaches.

\textbf{LexRank}~\cite{erkan2004lexrank}, which is a popular general-purpose extractive summarizer, relies on a graph representation of the document(s) building a complete graph, where each sentence from the document set becomes a node, and each edge weight is defined by the value of the cosine similarity between the sentences. Then the centrality index of each node is computed producing a ranking of vertices based on their importance, which indicates the ranking of the most relevant sentences to compose the summary. 

% This well-known strategy is used as basis for many recent unsupervised approaches on automatic text summarization~\cite{tsur2009revrank,wu2011unsupervised,cheng2011relin,woloszyn2017mrr,al2017sentiment}

% Another example of the unsupervised technique is Al-Dhelaan's work~\cite{al2017sentiment} that performs contrastive opinion summarization, which provides two opposing points of views about the same product. It builds a relation graph of reviews for each sentiment (positive and negative) and chooses the salient ones from each class using PageRank.

% Nevertheless, these approaches do not take into consideration the repetition of words problem that causes redundancy. Neither they present a conceptual model to meet the cross-domain summarization demands. Thus, next Section describes an unsupervised cross-domain summarization model and a post-processing algorithm to reduce repetition and improve the coverage of summaries.

This well-known strategy is used as the basis for many recent unsupervised approaches~\cite{woloszyn2018distrustrank,dos2018ddc,woloszyn2017mrr,WoloszynBeatnik2017}. Nevertheless, these approaches do not take into consideration the repetition problem that causes redundancy of words. Neither they present a conceptual model to meet the cross-domain summarization demands. Thus, next Section describes an unsupervised cross-domain summarization model and a post-processing algorithm to reduce repetition and improve the coverage of summaries.

%cite old wu2011unsupervised,cheng2011relin,woloszyn2017mrr,al2017sentiment
%
%
%
\section{Developed Model}\label{sec:model}
The developed model structures a given text set in a graph model, and uses another specialized text set, from another domain, to put a bias in the extracted summary. As already described sometimes it is necessary to extract a specialized summary from a more general-purpose text set. The example given before is related to extracting educational aspects from user comments in movies. Besides the domain bias, the model also structures a post-processing that treats the problem of sentences repetition.

\begin{figure}[t!]
     \centering
     \includegraphics[width=1\textwidth]{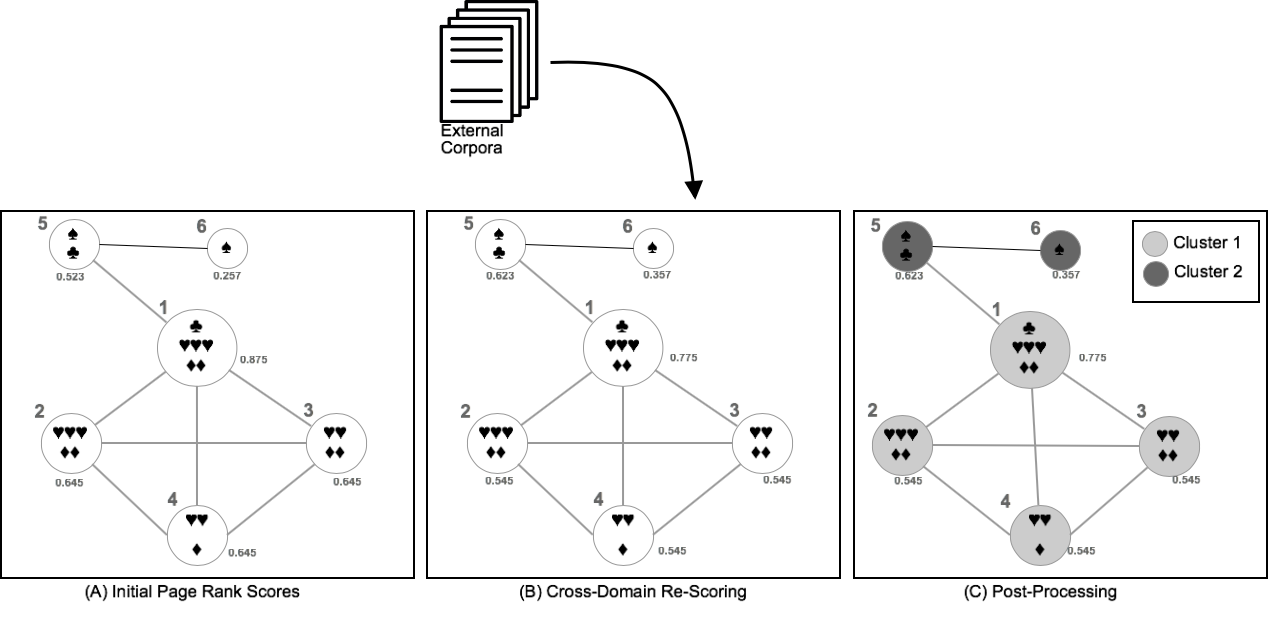}
 \caption{Illustration of the cross-domain Graph centrality building, and the post-processing to avoid redundancy.}
 \label{fig:Prop_appr}
 \end{figure}

In Figure~\ref{fig:Prop_appr}, it is shown how the cross-domain redundancy-free summary is extracted by using the Graph model. Since the first steps are the same of a general Graph-based summary (shown in Figure~\ref{fig:Repe_GC}), this process starts with the output of the general process, i.e., a Graph where each node have a Page-Rank score that represents how central a determined sentence is (Figure~\ref{fig:Prop_appr}(A)).

The initial Page-Rank scores are then recomputed by taking in consideration keywords found on an external corpus. Such keywords are used as a bias to compute the importance of each sentence. The final specialized summary is based on the centrality score of the sentences weighted by the presence of keywords from the external corpora.

%beatnik(beatnik algorithm)
Let $S$ be a set of all sentences extracted from the $R$ user's reviews about a single movie; the first step is to build a graph representation $G=(V,E)$, where $V=S$ and $E$ is a set of edges that connect pairs $\langle u,v \rangle \in V$. The score of each node (that represent a sentence) is given by the harmonic mean between its centrality score on the graph given by PageRank, and the sum of the frequencies of its specialized keywords (stated in equation \ref{similarity_socore_}). The pseudo-code of the Cross-domain Re-scoring is given in Algorithm \ref{mri-algo}, where $G$ is represented as the adjacency matrix $W$.

{\centering
\begin{minipage}{.8\linewidth}
\begin{algorithm}[H]
\caption{ - Cross-domain Re-scoring Algorithm (${S}$,${B}$): ${O}$}
\label{mri-algo}
- Input: a set of sentences extracted from a general purpose corpora (e.g., Amazon's movies Reviews) $R$, and a corpora $B$ used as a bias \\
- Output: an extractive biased summary $O$ based on the general purpose corpora $R$. \\

\begin{algorithmic}[1]
\For {each $u,v$ $\in$ ${S}$}
\State $W[u,v]\leftarrow$ {\small \textit{idf-modified-cosine}(u,v)}
\EndFor
\For {each $u,v$ $\in$ $S$}
\If {$W[u,v]$ $\geq \beta$}
\State $W'[u,v]\leftarrow1$
\Else
\State $W'[u,v]\leftarrow0$
\EndIf
\EndFor
\State $P \leftarrow PageRank(W')$
\For {each $u$ $\in$ ${S}$}
\State $K\leftarrow$ {\small \textit{sim-keyword}(u, $B$)}
\State $O[u] \leftarrow \frac{ \|S\| P_uK}{P_u+K} $
\EndFor
%\State $O \leftarrow \frac{ \|S\| P K}{P+K} $
\State return $O$
% * <lwives@gmail.com> 2018-04-10T01:36:58.586Z:
% 
% > \State return $O$
% Esse return está no lugar certo? não seria fora do for mais externo? Alinhado aqui parece que está dentro (e cada passo do for iniciado na linha 4 retornaria algo? 
% 
% ^.
\end{algorithmic}
\end{algorithm}
\end{minipage}
\par
}

The main steps of the Cross-domain Re-scoring algorithm are: (a) it builds a similarity graph ($W$) between pairs of texts of the same product or subject (lines: 1-3); (b) the graph is pruned (W') by removing all edges that do not meet a minimum similarity threshold, given by the parameter $\beta$\footnote{The best parameter obtained in our experiments is $\beta=0.1$ } (lines 4-10); (c) using PageRank, the centrality scores of each node is calculated (line 11); (d) using the educational corpora, each sentence is scored according the presence of educational keywords (line 13); (e) The final importance score of each node is given by the harmonic mean between its centrality score on the graph, and the sum of its education keywords frequencies (line 14). 

To get the similarity between the two nodes we define an adapted metric, that is the cosine difference between two corresponding sentence vectors \cite{erkan2004lexrank}:
\begin{equation}
\textrm{idf-modified-cosine}(x,y) = 
\frac{\sum_{w\in{}x,y}
\textrm{tf}_{w,x}\textrm{tf}_{w,y}(\textrm{idf}_w)^2}
{\sqrt{\sum_{x_i\in{}x}(\textrm{tf}_{x_i,x}\textrm{idf}_{x_i})^2} \times
\sqrt{\sum_{y_i\in{}y}(\textrm{tf}_{y_i,y}\textrm{idf}_{y_i})^2}}
\end{equation}

where  $\textrm{tf}_{w,s}$ is the number of occurrences of the word $w$ in the sentence $s$. We employed the approach described by \cite{mihalcea2004textrank} to extract the keywords from the external corpora. The similarity between the sentences and the keywords extracted from the external corpora are given by the following equation:
\begin{equation}
\label{similarity_socore_}
\textrm{sim-keyword}(x,B) = 
\sum_{w\in{}x}
\textrm{tf}_{w \in keywords(B)}
\end{equation}

To reduce the textual redundancy problem, it was developed an algorithm that employs a clustering technique to find groups of sentences from the graph of sentences that are both homogeneous and well separated, where entities within the same group should be similar and entities in different groups dissimilar. Then, it takes the most central sentence from each group to compose the final summary. While  Graph-Centrality chooses the sentences based on their centrality, our algorithm divides the graph into \textit{k} groups of sentences and chose the most central sentence from each group Figure \ref{fig:Prop_appr}(C).

In the literature we find work employing clustering paradigms to provide a non-redundant multi-view of textual data~\cite{yang2010online,zhai2011clustering}. The agglomerative hierarchical clustering method is one of them, and it creates a hierarchy tree, or Dendrogram, which can be used for sentence coverage searching purposes. Conceptually, the process of agglomerating documents creates a cluster hierarchy for which the leaf nodes correspond to individual sentences, and the internal nodes correspond to merged groups of clusters. When two groups are merged, a new node is created in this hierarchy tree corresponding to this bigger merged group. %%gui: Por que falar desse método se usa outro?

In our work, we employed the Complete Link Hierarchical clustering algorithm~\cite{aggarwal2012mining} since it achieves better results on the experiments carried out, when compared with other clustering techniques, such as $k$-Means, $k$-Medoids, and EM~\cite{yang2010online,aggarwal2012mining}. By default, we remove stop words, and the remaining terms of the sentence are represented as uni-grams weighted by the known Term Frequency-Inverse Document Frequency (TF-IDF).

The pseudo-code for decreasing redundancy is displayed in Algorithm \ref{capo-algo}, where $G$ represents a complete graph obtained from the ATS approach based on Graph Centrality and cross-domain Re-scoring \ref{fig:Prop_appr}(B). $L$ represents the cluster labels extracted using the function \textit{cluster(G)} and $S$ the final solution containing \textit{k} sentences. 

%\begin{minipage}{.8\linewidth}
\begin{algorithm}[h!]
    \caption{ - Post-Processing Redundancy  Algorithm ($\mathscr{G}$, $\mathscr{P}$,$\mathscr{K}$): $\mathscr{S}$}
    \label{capo-algo}
    - Input: a complete graph $G=(V,E)$, where $V$ are the sentences and $E$ is a set of edges that represents the similarity between sentences; $\mathscr{P}$ the centrality score of each node; $K$ number the sentences to extract. \\
    - Output: ordered list $\mathscr{S}$ of sentences.  \\
    
    \begin{algorithmic}[1]
        \State $\mathscr{S}\leftarrow$ {\{\}}
        \State $\mathscr{L}\leftarrow$ {$cluster(G,K)$}
         
        \For {each $k$ $\in$ $\mathscr{K}$}
            \State $\mathscr{C_k}\leftarrow$ {$L[k]$}
            \State $\mathscr{CS_{k}}\leftarrow$ {$sort\_nodes\_by\_centrality(C_k,P)$}
             \State $\mathscr{S[k]}\leftarrow$ {$CS_{k}[0]$}
        \EndFor
        \State return $\mathscr{S}$
    \end{algorithmic}
\end{algorithm}
%\end{minipage}
%
%
%
%

\section{Case Study}\label{sec:case_study}
%
%lwives - focar em como criar um grafo para determinada área e mostrar que o grafo permite compreender o conteúdo do sumário de maneira rápida e fácil (melhor do que um texto). Não importa a área... Mas depois mostrar que se for com vies, fica melhor...
As explained before, the goal of the presented approach was to build a graph representation of the main concepts of a document or set of documents. This representation works as a cross-domain summary avoiding redundancy. To do so, we selected two application domains, one to validate the cross-domain summary generation and other to validate the redundancy. The cross-domain generation was tested in the educational domain, and the redundancy control was tested in the news domain. Three datasets were employed to perform the experiments. %% gui:verificar se ficarão três mesmo ou se vai unificar os experimentos
% * <lwives@gmail.com> 2018-04-10T01:43:21.927Z:
% 
% > the goal of the presented approach was to build a cross-domain summary avoiding redundancy.
% Mas o objetivo, até pelo título, seria criar um grafo representativo. Entendo que evitar redundância seja ok, mas não estava legal. Mudei um pouco.
% 
% ^ <lwives@gmail.com> 2018-04-10T01:46:30.555Z.

The first served as a word thesaurus to implement the educational bias in the cross-domain generation, and it was collected from an educational website~\footnote{http://www.teachwithmovies.org/index.html} TeachWithMovies (TWM) where a set of movies are described by teachers with the goal to use them as learning objects inside a classroom. The second dataset is Amazon Movie Reviews (AMR) \cite{McAuley:2013:ACM:2488388.2488466} which provides user comments about a large set of movies. Since we were interested in movies that appeared in both datasets, a filter was applied, and we ended up with 256 movies to perform our evaluation.

The third was used to evaluate the post-processing that which treats the redundancy problem. CSTNews\footnote{public available on http://conteudo.icmc.usp.br/pessoas/taspardo/sucinto/cstnews.html}~\cite{cardoso2011cstnews} is a novel corpus which comprises 140 news texts in Brazilian Portuguese divided into 50 groups, which has been successfully employed as gold-standard for many recent works on content selection and automatic production of summaries~\cite{cardoso2015exploring,maziero2015semi,dias2015discursive,nobrega2016improving,condori2017opinion}. Next, we describe each dataset with more details.

\subsection{Teaching with Movies}
The TeachWithMovies dataset was collected through a crawler developed by us. Different teachers described the movies on the website, but each movie has only one description, this was a challenge while collecting the data because the information was not standardized or had associated metadata.

However, we have noticed that some movies presented common information such as: i) movie description; ii) rationale for using the movie; iii) movie benefits for teaching a subject; iv) movie problems and warnings for young watchers; and v) objectives of using this movie in class. The developed crawler extracted such information, and we have used the movie description since it contains the greatest amount of educational aspects. In the end, 408 unique movies and video clips were extracted, but after matching with the Amazon dataset, we could use 256 movies. This dataset was used as a Gold-standard to cross-domain summary generation.

\subsection{Amazon Movie Reviews}
The Amazon Movie Reviews was collected with a timespan of more than ten years and consists of proximately 8 millions of reviews that include product and user information, ratings, and a plain text review.In Table~\ref{T1} is shown some statistics about the data.

\begin{table}[h]
\centering
\caption{Amazon Movie Reviews Statistics}
\label{T1}
\begin{tabular}{lr}
\hline
\multicolumn{2}{c}{Dataset Statistics}                   \\ \hline
Number of reviews                  & 7,911,684           \\
Number of users                    & 889,176             \\
Expert users (with \textgreater 50 reviews) & 16,341      \\
Number of movies                 & 253,059             \\
Mean number of words per review    & 101                 \\
Timespan                           & Aug 1997 - Oct 2012 \\ \hline
\end{tabular}
\end{table}

\subsection{CSTNews}

In this dataset, each group of news has from 2 to 3 texts on the same topic, having in average 49 sentences and 945 words. It comprises clusters of news texts manually annotated in different ways to discursive organization, Rhetorical Structure Theory and Cross-document Structure Theory annotations. The corpus includes manual multi-document summaries (one for each cluster of news) with 70\% compression rate (in relation to the longest text). The texts are manually annotated with high-level of agreement (more than 80\%) of human judges using Cohen's kappa coefficient, which is a statistic to measure inter-rater agreement for classification tasks~\cite{carletta1996assessing}. That means the annotation agreement is reliable and similar to that in presented in other works \cite{da2011development} for other languages than Portuguese. For such reason, these human-generated summaries were used as a Gold-standard. Since this post-processing is not language dependent and the redundancy is also a problem observed in different languages~\cite{saggion2013automatic}, this corpus was used to evaluate this post-processing strategy. 
\section{Experiment Design}
\label{sec:experiment_design}
This section presents the experimental setting used to evaluate the cross-domain summary generation and the post-processing that reduces redundancy. It describes the methods employed as the baselines for comparison, the educational plans adopted as Gold-standard and the metrics applied for evaluation, as well as details of the experiment, performed to assess the approach.

\subsection{First Baseline}

The results obtained from our cross-domain summary are compared with Textrank~\cite{mihalcea2004textrank} algorithm. Textrank was chosen because it is also a graph-based ranking algorithm and has been widely employed in Natural Language tools. Textrank essentially decides the importance of a sentence based on the idea of “voting” or “recommending”. Considering that in this approach each edge represents a vote, the higher the number of votes that are cast for a node, the higher the importance of the node (or sentence) in the graph. The most important sentences compose the final summary.

\subsection{Second Baseline}

Centrality-based ranking has been successfully on recent works to content selection and automatic production of textual summaries~\cite{wu2011unsupervised,cheng2011relin,thomas2015exb,woloszyn2017mrr,al2017sentiment}. LexRank~\cite{erkan2004lexrank} is a well-know ATS system based on Graph Centrality that has been used many times in the literature for comparisons purposes, due to its good performance. Since the post-processing strategy aims to reduce redundancy in Centrality-based approaches, we employ LexRank as baseline due it uses only the sentence centrality index to the ranking task. We used MEAD's implementation of Lexrank \cite{radev2004mead}, which is a publicly available (for researching purposes) framework for text summarization that provides a set of Perl components for the summarization of texts written in English as well as in other languages such as Chinese.

\subsection{Evaluation Metrics}
\subsubsection{ROUGE-\textit{n}}
The evaluation was performed by applying ROUGE (Recall-Oriented Understudy for Gisting Evaluation)~\cite{lin2004rouge}, which is a metric inspired on the Bilingual Evaluation Understudy (BLEU) \cite{saggion2013automatic}. Specifically, we used ROUGE-\textit{n} in the evaluation, this version of ROUGE makes a comparison of n-grams between the summary to be evaluated and the ``gold-standard"; in our case, cross-domain summaries and TWM lesson plans, respectively. We evaluated the first 100 words of the summaries obtained by our approach and the baseline since it corresponds to the median size of the gold-standard. ROUGE was chosen because it is one of most used measures in the fields of Machine Translation and Automatic Text Summarization~\cite{poibeau2012multi}.

\subsubsection{Redundancy}
An important aspect related to redundancy is lexical cohesion. Therefore, cohesive links between sentences is a positive component of the summary, and it has long been considered a key component in assessing content relevance in text summarization~\cite{saggion2013automatic}. However, in some cases, it could improve mean redundancy in summaries. To show how redundancy would affect a multi-document summarizer, we perform a comparison between the baseline and human gold-standard summary. 

\subsubsection{Coverage}
It is the extent to which all words of the automatic summaries are found in the source documents. In other words, it is a global score assessing to what extent the candidate summary covers the text given as input. 
%This metric is commonly explored in other areas, such as database systems~\cite{das2011mri,drosou2012disc}.
%
%
%
%

\section{Results}\label{sec:results}
The next subsections present the evaluation results.

%beatnik(Results)
\subsection{Cross-domain Summaries}
In this section, we present cross-domain summaries evaluation regarding the adopted baselines concerning precision, recall, and f-Score obtained by using ROUGE-N.

The gold-standard utilized in the experiments, as already stated is the educational description extracted from the TWM website. Table~\ref{T_results} shows the mean Precision, Recall, and F-Score, considering both our cross-domain strategy and Textrank (the gold-standard used as the baseline).

The results presented in Table~\ref{T_results} show that our strategy outperformed the baseline in all measurements carried out. Regarding Precision, the differences range from 4.9 to 11.9 percentage points (pp) on all ROUGE-N analyzed, where N is the size of the n-gram used by ROUGE. Using Wilcoxon statistical test with a significance level of 0.05, we verified that our strategy is statistically superior when compared to the baseline. Regarding recall, the differences are also in favor of our strategy, ranging from 4.7 to 11.5 pp when compared to the baseline.

\begin{table}
\centering
\caption{Mean of ROUGE results achieved by the Baseline (Column A) and our cross-domain strategy (Column B)}
\label{T_results}
\begin{tabular}{llll}
\\
\hline
\textit{ROUGE-n} & \textit{Column A} & \textit{Column B} & \textit{p}-values         \\ \hline
\textit{Precision-1}                 & 0.65615   &  \textbf{0.77028 }   & $<$ 0.05  \\
\textit{Recall-1}                    & 0.65003   &  \textbf{0.75611 }   & $<$ 0.05  \\
\textit{F\_score-1} 			 	 & 0.65283   &  \textbf{0.76296 }& $<$ 0.05     \\ \hline
\textit{Precision-2}                 & 0.22394   &  \textbf{0.34350  }  & $<$ 0.05  \\
\textit{Recall-2}                    & 0.22192   &  \textbf{0.33744 }   & $<$ 0.05  \\
\textit{F\_score-2} 				 & 0.22284   &  \textbf{0.34037 }& $<$ 0.05     \\ \hline
\textit{Precision-3}                 & 0.06313   &  \textbf{0.11268 }   & $<$ 0.05  \\
\textit{Recall-3}                    & 0.06387   &  \textbf{0.11102 }   & $<$ 0.05  \\
\textit{F\_score-3} 				 & 0.06347   &  \textbf{0.11182 }& $<$ 0.05     \\
\hline
\end{tabular}
\end{table}

\begin{figure*}[h!]
    \centering
    \includegraphics[width=0.7\textwidth]{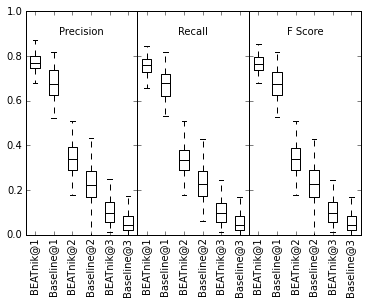}
\caption{Distribution of Rouge results.}
\label{fig:boxplo}
\end{figure*}

Regarding the distribution of Rouge's results, in Fig \ref{fig:boxplo} it is shown a boxplot indicating that our strategy results are not only better in mean, but also in terms of lower and upper quartiles, minimum and maximal values. 

%

%capoeira(Results and Discussion)
\subsection{Post-processing}
In this section, we will discuss the results obtained in our experiments regarding the adopted baselines in terms of Coverage, Redundancy, Precision, and Recall using CSTNews.

\subsubsection{Redundancy and Coverage}
\begin{figure}[!htb]
   \begin{minipage}{0.49\textwidth}
     \centering
     \includegraphics[width=1\linewidth]{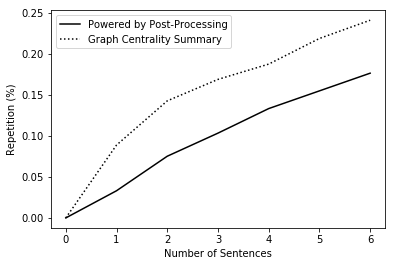}
     \caption{Mean Redundancy}\label{Fig:repetition}
   \end{minipage}\hfill
   \begin {minipage}{0.49\textwidth}
     \centering
     \includegraphics[width=1\linewidth]{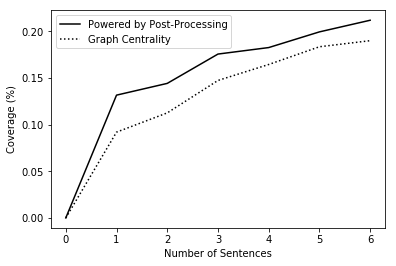}
     \caption{Mean Coverage}\label{Fig:coverage}
   \end{minipage}
\end{figure}

Figures \ref{Fig:repetition} and \ref{Fig:coverage} show that our post-processing strategy outperformed the unsupervised baseline generation summaries with less redundancy and more coverage, being closer to the human gold-standard summaries. The mean redundancy differences range from 5.42 to 6.75 percentage points (pp) when compared to Lexrank. In terms of coverage, the mean difference is up to 3.96 pp. Using a Wilcoxon statistical test with a significance level of 0.05, we verified that our strategy results are statistically superior both in redundancy and coverage.

\subsubsection{Precision and Recall}
\begin{figure}[!htb]
   \begin{minipage}{0.49\textwidth}
     \centering
     \includegraphics[width=1\linewidth]{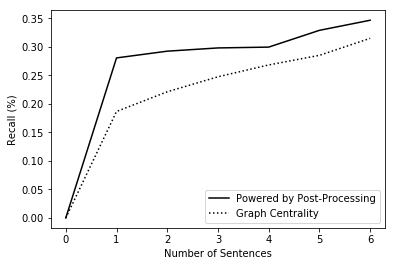}
     \caption{Mean Recall}\label{Fig:recall}
   \end{minipage}\hfill
   \begin {minipage}{0.49\textwidth}
     \centering
     \includegraphics[width=1\linewidth]{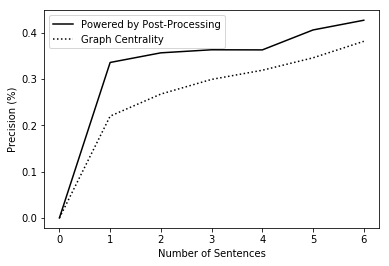}
     \caption{Mean Precision}\label{Fig:precision}
   \end{minipage}
\end{figure}

Figure \ref{Fig:recall} and \ref{Fig:precision} show that our strategy also outperformed the unsupervised baseline in terms of Recall and Precision obtained using ROUGE-1. For Recall, the mean differences ranging from 3.12 to 9.39 pp when compared to Lexrank. For Precision, the mean differences ranging from 4.39 to 11.57 pp in all cases. Using the Wilcoxon statistical test with a significance level of 0.05, we verified that our strategy results are statistically superior in all cases.

\section{Conclusion}\label{sec:conclusion}
%beatnik
In this paper, we presented an approach to generate cross-domain summaries based on graphs that are able to represent the main concepts of a document or set of documents. The proposed approach also reduces text redundancy in the generated summaries. We showed that our approach achieved statistically superior results than Textrank (a general summary algorithm) and Lexrank (another general summary algorithm).

The proposed algorithms require no training data, which avoids costly and error-prone manual training annotations. Compared to the baselines, our approach: a) outperforms the unsupervised techniques in terms of Precision and Recall; b) Statistically reduces redundancy and improves coverage; and c) is easy to plug into any standard Graph Centrality approach in any domain. Our experiments were performed in two domains, the educational and the news one, attesting the approach versatility.

%{\color{red} É preciso um parágrafo mostrando que pode ser aplicado a outros problemas} 
%%gui: deixei o parágrafo anterior mais genérico, mostrando que utilizamos o approach em 2 domínios distintos.

Finally, it is also important to state that we found out a considerable number of highly helpful sentences with low centrality indexes which lead us to consider the investigation of other techniques to select the most relevant sentences to compose the movies' educational description. It is also important to reaffirm the approach source language independence, for that reason, we consider, in the future, to extend the evaluation using different languages and summaries length.

%capoeira
% In this work, we propose CaPOEIrA, a novel unsupervised post-processing algorithm that reduces redundancy and improves coverage of summaries based on a clustering technique. The intuition behind this approach is that choosing the most \textit{k}-dissimilar sentences from the weighted graphs cluster would reduce repetition, which is an issue in ATS, while improving the coverage of the final summary.

% The proposed algorithm requires no prior domain knowledge and no training data, which avoids costly and error-prone manual training annotations. Compared to the baseline, CaPOEIrA: a) outperforms the unsupervised technique in terms of Precision and Recall; b) Statistically reduces redundancy and improves coverage; and c) is easy to plug into any standard Graph Centrality approach.

% Despite CaPOEIrA is independent of the source language, as a future work, we consider extending the evaluation using different languages and summaries length. We also consider performing an evaluation in noisy domains, such as summarization of the product reviews on the internet.

\bibliographystyle{splncs04}
\bibliography{refs.bib}

\end{document}